\def\year{2021}\relax
\documentclass[letterpaper]{article} 
\usepackage{aaai21}  
\usepackage{times}  
\usepackage{helvet} 
\usepackage{courier}  
\usepackage[hyphens]{url}  
\usepackage{graphicx} 

\usepackage[utf8]{inputenc} 
\usepackage[T1]{fontenc}    
\usepackage{hyperref}       
\usepackage{booktabs}       
\usepackage{amsfonts}       
\usepackage{nicefrac}       
\usepackage{microtype}      

\usepackage{amsmath}
\usepackage{multirow}
\usepackage{caption}
\usepackage{subcaption}
\usepackage{booktabs}
\usepackage{enumitem}
\usepackage{mathtools}
\usepackage{natbib}  
\usepackage{caption} 
\title{Show, Attend and Distill: \\ Knowledge Distillation via Attention-based Feature Matching}
\author {
    Mingi Ji \textsuperscript{\rm 1}\footnote{This work was done at internship in CLOVA AI Research, NAVER Corp.},
    Byeongho Heo \textsuperscript{\rm 2},
    Sungrae Park \textsuperscript{\rm 3} \\
}
\affiliations {
    \textsuperscript{\rm 1} Korea Advenced Institute of Science and Technology (KAIST) \\
    \textsuperscript{\rm 2} NAVER AI LAB \\
    \textsuperscript{\rm 3} CLOVA AI Research, NAVER Corp. \\
    qwertgfdcvb@kaist.ac.kr, bh.heo@navercorp.com, sungrae.park@navercorp.com
}

\begin{document}

\maketitle
\begin{abstract}
Knowledge distillation extracts general knowledge from a pre-trained teacher network and provides guidance to a target student network. Most studies manually tie intermediate features of the teacher and student, and transfer knowledge through pre-defined links. However, manual selection often constructs ineffective links that limit the improvement from the distillation. There has been an attempt to address the problem, but it is still challenging to identify effective links under practical scenarios. In this paper, we introduce an effective and efficient feature distillation method utilizing all the feature levels of the teacher without manually selecting the links. Specifically, our method utilizes an attention-based meta-network that learns relative similarities between features, and applies identified similarities to control distillation intensities of all possible pairs. As a result, our method determines competent links more efficiently than the previous approach and provides better performance on model compression and transfer learning tasks. Further qualitative analyses and ablative studies describe how our method contributes to better distillation. The implementation code is available at \textit{github.com/clovaai/attention-feature-distillation.}
\end{abstract}

\section{Introduction}
\begin{figure*}[t]
    \centering
    \includegraphics[width=1.9\columnwidth]{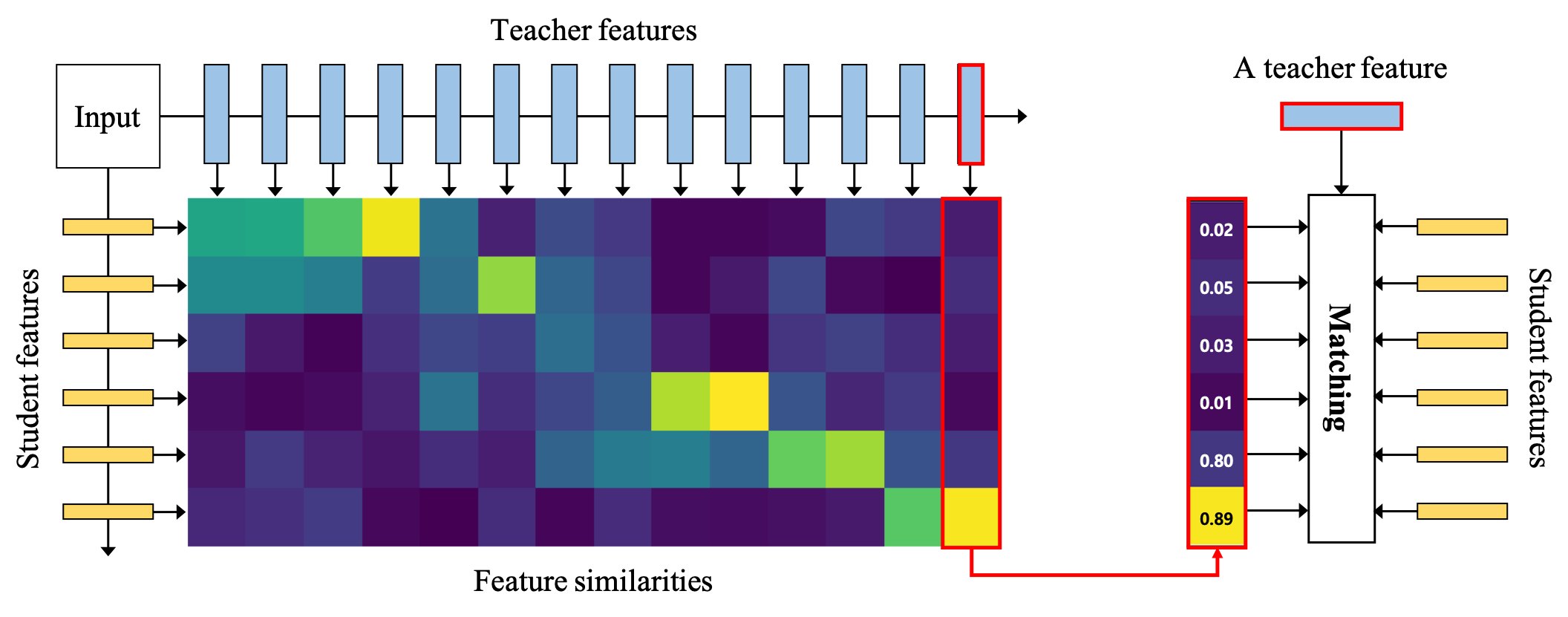}
    \caption{Overview of AFD. An attention-based model determines similarities between the teacher and student features. Knowledge from each teacher feature is transferred to the student with the identified similarities.}
    \label{fig:overview}
\end{figure*}

Knowledge distillation is the technique for transferring knowledge from a source neural network to a target neural network~\cite{hinton2015distilling}. The source network, referred to as \emph{a teacher}, indicates a large network that is highly regularized via pre-training, and the target network, referred to as \emph{a student}, is a smaller network for a specific task. 
The pre-trained teacher directly informs the student of the solution and intermediate process of a problem, and this informative supervision enables fast and effective learning of the student.
Based on knowledge distillation, recent studies have shown significant improvements in model compression~\cite{hinton2015distilling,fitnets,fsp,crd}, cross-domain transfer learning~\cite{kd_domain,large_domain,acoustic_domain}, and continual learning~\cite{lwf,lifelong}.

For the success of knowledge distillation, various distillation methods were introduced.
Starting from transferring output probability distributions of the teacher~\cite{hinton2015distilling}, intermediate features representations~\cite{fitnets} and their variants~\cite{atts,rkds,crd} are investigated to identify what knowledge of the teacher helps to build a better student.
However, most studies manually links the teacher and student features and perform distillation through the links individually.
This manual link selection does not consider the similarity between the teacher and student features, so there is a risk of forcing an incorrect intermediate process to the student.
Furthermore, the link selection has a limitation on fully utilizing the whole knowledge of the teacher by choosing a few of all possible links.

To compensate for the limitation, Jang \textit{et al.}~\cite{l2t} apply a meta-networks, ``learning to transfer (L2T)'', automatically determining the links.
In more details, the meta-network consists of individual gates for all possible links, and each gate determines whether distillation through the link contributes to decreasing the classification loss of the student.
Their results prove that knowledge distillation with the identified links provides better performance than those with manually selected links. 
However, the individual gates are not aware of each other although the distillation through the gates simultaneously affect the student. 
Moreover, their meta-learning scheme requires expensive inner-loop procedures to learn their meta-networks, thus its application can be limited under practical scenarios.

In this paper, we introduce a new feature linking method based on an attention mechanism~\cite{show_attend_tell,transformer}, which is called attention-based feature distillation (AFD).
Specifically, AFD utilizes an attention-based meta-network that identifies similar features between the teacher and student.
The identified similarities are applied to control the distillation intensities for  all possible feature pairs. 
Figure~\ref{fig:overview} shows an overview of our proposed distillation method to provide graphical descriptions. 

When comparing from L2T, our proposed method considers the granularity of the teacher and student features to identify the importance of their links while L2T only uses information for a single pair in a narrow perspective. In addition, AFD learns from feature similarities without any inner-loop procedure but L2T learns from the classification loss, which requires expensive Hessian computation. In our experiment, we observe that L2T and ours have distinct linking results from the different objectives, but our method shows better or comparable results on multiple tasks with more efficient computation.  

We conduct experiments for model compression on three image classification tasks such as CIFAR-100~\cite{cifar100}, tinyImageNet, and ImageNet~\cite{imagenet} and for domain transfer on four specific tasks such as CUB200~\cite{cub}, MIT67~\cite{mit67}, Stanford40~\cite{stan40}, and Stanford Dogs~\cite{stan_dog} with a pre-trained large network on ImageNet. As a result, our method shows performance gains in most of experiments and the analyses on the identified feature links explains how our method works. Further ablation studies and sensitivity analysis provide a guideline to use our method. 

\section{Related Work}
Hinton \textit{et al.} introduced the concept of knowledge distillation \cite{hinton2015distilling} by utilizing the output probability distributions of the teacher as a soft label to transfer knowledge. Also, intermediate features from the teacher have been proved to hold additional knowledge that can contribute to improving the student performance. Romero \textit{et al.} (FitNet~\cite{fitnets}) proposed feature-based distillation that couples the teacher and student features and induce the student to mimic the paired teacher features. However, due to a capacity gap between the teacher and the student, it is challenging for the student to mimic the exact teacher features.

To address the problem, recent works focused on propagating core knowledge from the teacher features. Zagoruyko \textit{et al.}~\cite{atts} simplified teacher features by applying channel-wise summations and led the student to learn core knowledge of refined teacher features. Kim \textit{et al.}~\cite{para} extracted low-dimensional representations of the features via multiple auto-encoders and transfer them to the student. Relational knowledge distillation aims to transfer relation knowledge between data instances~\cite{rkds,sp,cckd,irg}. In the other direction of feature refinement methods, the distillation regularization terms have been explored to allow the student to accept more knowledge. Ahn \textit{et al.}~\cite{vid} transferred knowledge by maximizing the mutual information between the feature of the teacher and student. Huang \textit{et al.}~\cite{nst} utilized the maximum mean discrepancy to propagate knowledge from the teacher features. Tian \textit{et al.}~\cite{crd} applied the contrastive learning scheme on relational knowledge distillation. 

Although distillation methods on how to refine and propagate knowledge have been continuously advanced as the above, it is still remaining problem how to link intermediate features between the teacher and student. Our method is placed to solve the problem as like L2T~\cite{l2t}. However, L2T and ours have different properties to identify the links.

\section{Attention-based Feature Distillation}
\begin{figure*}[t]
    \centering
    \includegraphics[width=1.9\columnwidth]{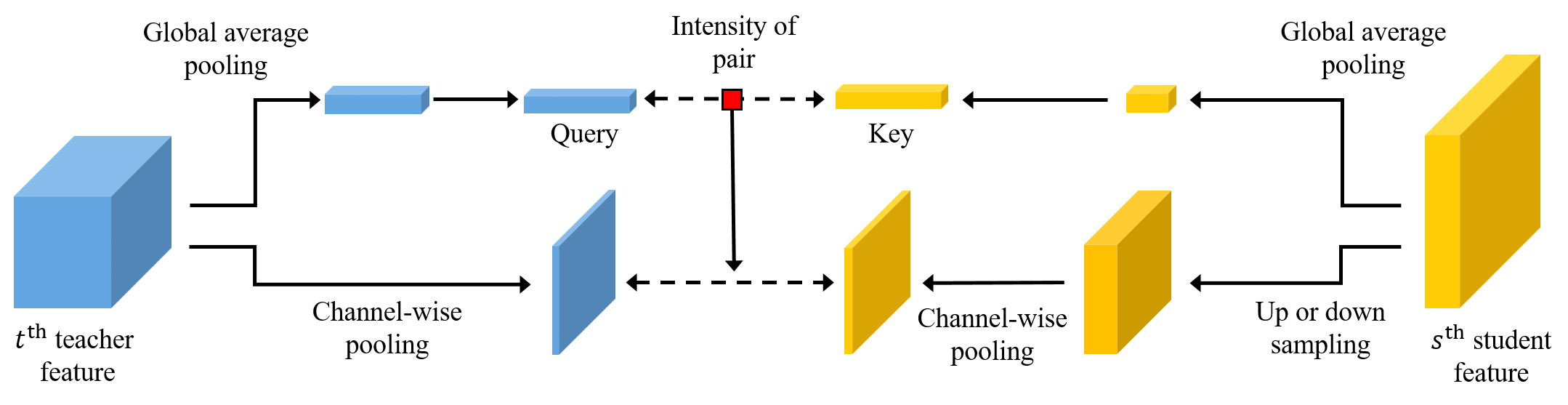}
    \caption{Overview of the proposed meta-network. The globally pooled features are utilized to estimate the similarities and the channel-wisely averaged features are used to calculate the distance between the features.}
    \label{fig:method}
\end{figure*}

Let $\mathbf{h}^{\text{T}}=\{{h}^{\text{T}}_1, ..., {h}^{\text{T}}_T\}$ be a set of the feature candidates from the teacher and $\mathbf{h}^{\text{S}}=\{{h}^{\text{S}}_1, ..., {h}^{\text{S}}_S\}$ be a set of feature candidates from the student where $T$ and $S$ indicate the numbers of the candidates from the teacher and student, respectively. Each candidate has its own feature map size and channel dimension as ${h} \in \mathbb{R}^{H \times W \times d }$ where $H$, $W$, and $d$ indicate the height, width, and channel dimension, respectively. When two sets of the candidates are given, AFD aims to identify similarities for all possible combinations ($S \times T$ pairs) and transfer knowledge of the teacher candidates to the student with the identified similarities. 

Figure~\ref{fig:method} shows the overview of our proposed network. As can be seen, the feature candidates are compared in two directions with the two pooling methods: global average pooling and channel-wise pooling. The similarity identified by two globally pooled features is used as an intensity for transferring knowledge through the distance defined by the channel-wisely averaged features. In order to identify the similarity between ${h}^{\text{T}}_t$ and ${h}^{\text{S}}_s$, AFD adopts a query-key concept of the attention mechanism~\cite{show_attend_tell,transformer}. Specifically, each teacher feature generates a query, $\mathbf{q}_t$, and each student feature identifies a key, $\mathbf{k}_{s}$. 
The followings describe $\mathbf{q}_t$ and $\mathbf{k}_{s}$ in mathematical expressions;
\begin{equation}
\begin{aligned}
    & \mathbf{q}_t = f_Q ( W^{\text{Q}}_t \cdot \phi^{HW}({h}^{\text{T}}_t) ), \\
    & \mathbf{k}_{s} = f_K ( W^{\text{K}}_s \cdot \phi^{HW}({h}^{\text{S}}_s) ).
\end{aligned}
\end{equation}
Here, $\phi^{HW}( \cdot )$ indicates a global average pooling. $f_Q$ and $f_K$ are activation function of the query and key. $W^{\text{Q}}_t \in \mathbb{R}^{d \times d^{\text{T}}_t}$ and $W^{\text{K}}_s \in \mathbb{R}^{d \times d^{\text{S}}_s}$ are linear transition parameters for the $t$-th query and the $s$-th key. It should be noted that the features have different transition weights since they have different properties through their different levels, \textit{i.e.} a low-level visual feature can represent a line and a high-level visual features can represent an object. Therefore, we apply different transition weights to each features.

By utilizing the queries and keys, attention values that represent relations between teacher and student candidates are calculated with a ``softmax'' function;
\begin{equation}
\begin{aligned}
    \mathbf{\alpha}_t={}&\text{softmax} ( [(\mathbf{q}_t^\top W_1^ {\text{Q-K}} \mathbf{k}_{t,1} + (\mathbf{p}_t^{\text{T}})^\top \mathbf{p}_{1}^{\text{S}}) / \sqrt{d}, \\
    & \cdots , (\mathbf{q}_t^\top W_S^{\text{Q-K}}\mathbf{k}_{t,S} + (\mathbf{p}_t^{\text{T}})^\top \mathbf{p}_{S}^{\text{S}}) / \sqrt{d} ]).
\end{aligned}
\end{equation}

Here, we introduce additional weight parameters; a bilinear weight, $W_t^{\text{Q-K}}\in \mathbb{R}^{d \times d}$, and positional encodings, $\mathbf{p}_{t}^{\text{T}} \in  \mathbb{R}^{d}$ and $\mathbf{p}_{s}^{\text{S}}\in  \mathbb{R}^{d}$. The bilinear weight is applied to generalize the attention value from different source ranks since the query and key are identified from different dimensional features~\cite{bilinear-1,bilinear-2}. The positional encodings are utilized to share common information over different instances~\cite{transformer}. $\mathbf{\alpha}_t$ is the attention vector that capture relation between the $t$-th teacher feature and whole student features. By utilizing $\mathbf{\alpha}_t$, the teacher feature, ${h}^{\text{T}}_t$, enables to transfer its knowledge selectively to student features.

The final distillation term forms as
\begin{equation}
    \mathcal{L}_{\text{AFD}} = \Sigma_{t}{\Sigma_{s}{ \alpha_{t,s} \left\Vert
    \tilde{\phi}^{C}({h}^{\text{T}}_t) - \tilde{\phi}^{C}(\hat{h}^{\text{S}}_s)
    \right\Vert_2 }},
    \label{eq:sad_loss}
\end{equation}
where $\tilde{\phi}^{C}$ indicates a combined function of a channel-wise average pooling layer with L2 normalization, $\mathbf{v}/\left\Vert \mathbf{v} \right\Vert_2$, by following \cite{atts}. In addition, $\hat{h}^{\text{S}}_s$ is up-sampled or down-sampled from $h^{\text{S}}_s$ to match the feature map size to those of the teacher features. 

Finally, the regularization term is added to the total loss function as following;
\begin{equation}
    \mathcal{L}_{\text{Student}} = \mathcal{L}_{\text{cls}} + \beta \mathcal{L}_{\text{AFD}},
\end{equation}
where $\mathcal{L}_{\text{cls}}$ is the classification loss with ground-truth labels and $\beta$ is a trade-off parameter controlling the impact of the proposed distillation loss. We use cross entropy for $\mathcal{L}_{\text{cls}}$. Using the loss, the student and the attention-based network are trained simultaneously. It should be noted that the AFD network is trained only with $\mathcal{L}_{\text{AFD}}$ that represents the weighted similarities for all possible feature pairs, so AFD does not require expensive Hessian computations to connect its parameters with $\mathcal{L}_{\text{cls}}$.

\section{Experiments}
\begin{table*}
\tabcolsep=0.17cm
\centering
\begin{tabular}{c|ccccc|cc}
\toprule
        & \multicolumn{5}{c|}{\textbf{Same style}}                                   & \multicolumn{2}{c}{\textbf{Different style}}           \\ \midrule
Teacher & \small{ResNet56}        & \small{ResNet110}      & \small{ResNet110}       & \small{WRN-40-2}        & \small{WRN-40-2}     & \small{WRN-40-2}            & \small{ResNet34}                   \\
Student & \small{ResNet20}        & \small{ResNet20}        & \small{ResNet56}        & \small{WRN-16-2}        & \small{WRN-40-2}     & \small{ResNet56}            & \small{WRN-28-2}                   \\ \midrule
Teacher & 0.7254          & 0.7409          & 0.7409          & 0.7620          & 0.7620           & 0.7620              & 0.7860                       \\
Student & 0.6940          & 0.6940          & 0.7254          & 0.7289          & 0.7620           & 0.7254              & 0.7532                       \\ \midrule
KD      & 0.7098          & 0.7081          & 0.7483          & 0.7499          & 0.7763           & 0.7497              & 0.7648                       \\
FitNet  & 0.7005          & 0.7002          & 0.7411          & 0.7522          & 0.7766           & 0.7506              & 0.7644                       \\
ATT     & 0.7054          & 0.7081          & 0.7488          & 0.7520          & 0.7778           & 0.7516              & 0.7720                       \\
RKD     & 0.7043          & 0.7076          & 0.7477          & 0.7459          & 0.7762           & 0.7439              & 0.7632                       \\
CRD     & 0.7095          & 0.7091          & 0.7512          & 0.7515          & 0.7780           & 0.7525              & 0.7697                       \\
L2T     & 0.7037          & 0.7001          & 0.7457          & 0.7486          & 0.7678           & 0.7463              & 0.7640                       \\ \midrule
Ours    & \textbf{0.7153} & \textbf{0.7138} & \textbf{0.7539} & \textbf{0.7547} & \textbf{0.7813} & \textbf{0.7540}     & \textbf{0.7747}     \\ \bottomrule
\end{tabular}
\vspace{0.3em}
\caption{Performance comparison on CIFAR-100. The teachers and the students have same or different architectural style. All experiments are repeated 5 times.}
\label{table:cifar}
\end{table*}

We evaluate our proposed method on model compression tasks that train a smaller, or better model under a limitation of the model capacity and transfer learning tasks that obtain a better model in a specific domain by utilizing a pre-trained model. Following by the quantitative evaluations, we qualitatively analyze how our method works. Finally we provide ablation studies to provide further properties of our methods. 

In our experiment, our baseline methods are a traditional knowledge distillation method (KD) firstly introduced by Hinton \textit{et al.}~\cite{hinton2015distilling}, three popular feature-level distillation methods (FitNet~\cite{fitnets}, ATT~\cite{atts}, CRD~\cite{crd}) that require a manual feature matching, and one feature-level distillation method (L2T) automatically linking the teacher and student features. For CRD, we set the number of negative samples same as the batch size of each experiment. Note that KD is applied to all baselines to reveal additional gains from the feature distillation methods. 

\subsection{Model Compression}

We demonstrate the effectiveness of the proposed distillation method on model compression tasks. The experiments are conducted on three popular benchmark datasets such as \textit{CIFAR-100} \cite{cifar100}, \textit{tinyImageNet}, and \textit{ImageNet}~\cite{imagenet}. We utilize Residual Network (ResNet)~\cite{resnet} and Wide Residual Network (WRN)~\cite{wrn} architectural styles. First, we conduct an experiment on CIFAR-100 that consists of $32\times 32$ sized color images for 100 object classes and has 50K training and 10K validation images. For data augmentation, the horizontal flipping and random cropping are applied. We set the batch size as 64 and the maximum iteration as 240 epochs. All models are trained with stochastic gradient descent with $0.9$ of momentum, weight decay as $5\times 10^{-4}$, initial learning rate as 0.05, and divide it by 10 at 150, 180, 210 epochs. For the baseline methods, we use their official code and their hyper-parameters. For the distillation loss of our model, we apply \{30, 50, 100, 200\} of beta, $\beta$, and choose the best performer. We provide how the performance changes depending on the value of $\beta$ in \textit{Sensitivity Analysis} section. For the feature candidates of our method, we use all output features of the teacher and student residual blocks as candidate for all experiments, except ResNet110. 
For ResNet110, we skip one for every two residual blocks and used only half of the entire residual blocks.
In detail, the number of the candidates become 9 for ResNet20, 27 for ResNet56, 27 for ResNet110, 8 for ResNet18, 16 for ResNet34, 6 for WRN-16-2, 12 for WRN-28-2, and 18 for WRN-40-2.

Table~\ref{table:cifar} shows our experiment settings and results on the CIFAR-100 dataset with various network architecture. We pre-train large teacher networks and utilize them to train smaller or same-scaled student networks. The experiments are divided into two groups according to the architectural style of the teacher and student. When considering the student without knowledge distillation, all students shows the worst performance over all experiment settings. With manually linked feature pairs, the baseline models including KD, FitNet, ATT, RKD and CRD show better performances than the vanilla students. 
When applying L2T that identifies feature links with individual gates, we observed worse performance than other baseline methods even though it identifies beneficial feature pairs to meet their objective. 
The reason of the degradation is that L2T tends to propagate high-level knowledge of the teacher to the low-level feature of the student (See \textit{Qualitative Studies on Feature Attention} section).
Intuitively, the low-level of a small network cannot mimic the high-level of a large network, which adversely affects performance.
Our method that utilizes an attention mechanism to identify similar features between the teacher and student shows the best performance over all experiment settings.
In particular, our method shows an improvement over ATT which uses the same feature distance for distillation. In other words, the proposed linking method significantly contributes to distillation performance.

In order to validate our method on more real world environment, we compare our method from other baseline methods on tinyImageNet and ImageNet~\cite{imagenet}. The tinyImageNet dataset consists of $64\times 64$ sized 100K training and 10K validation images for 200 object classes and the ImageNet dataset includes 1.2M training and 50K validation large-scale images for 1K object classes. 
\begin{figure*}[t]
    \centering
    \includegraphics[width=1.9\columnwidth]{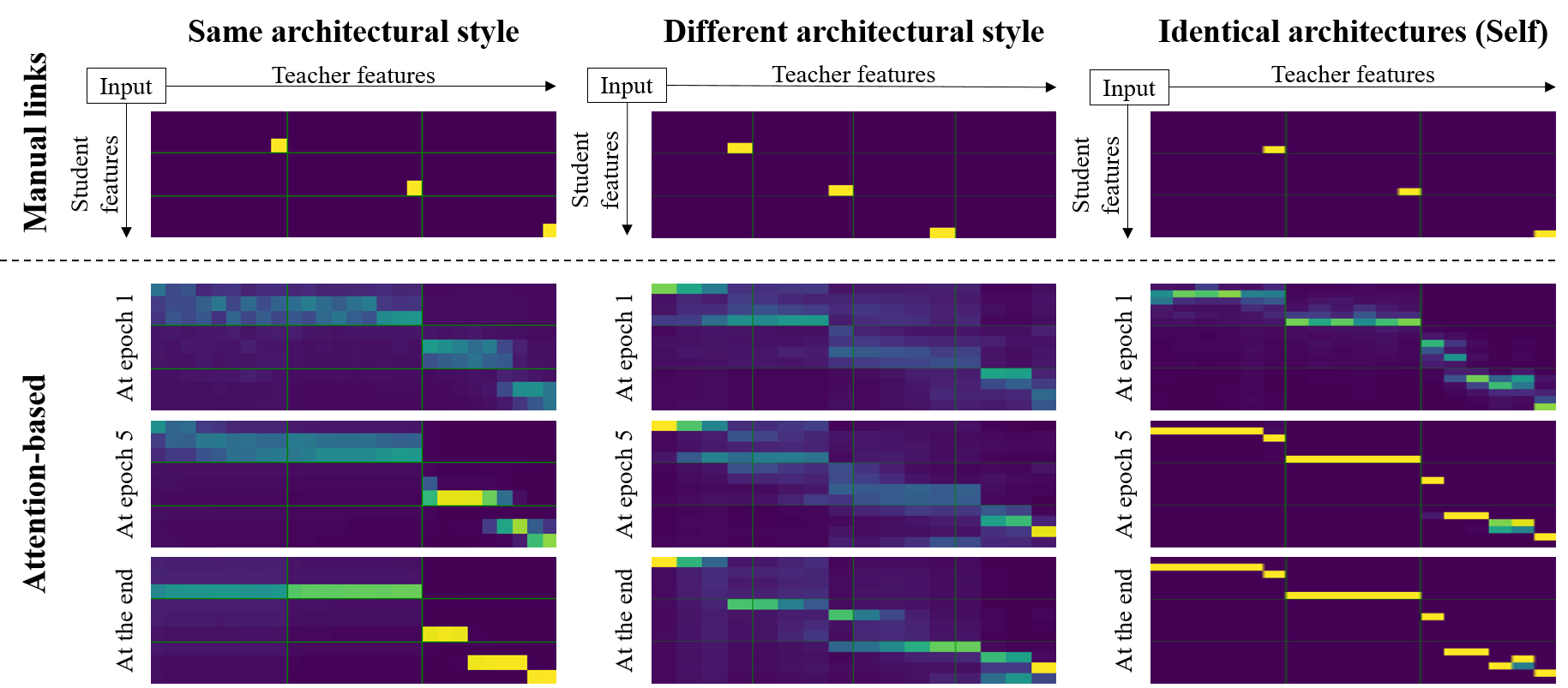}
    \caption{
    Manual and attention-based feature links for knowledge distillation. Rows and columns of matrices indicate the student and teacher features, respectively. Each matrix is the average overall $\alpha$ at the corresponding training epoch. The pairs are compared on multiple distillation settings; ResNet56 $\rightarrow$ ResNet20 (\textbf{Same architectural style}), ResNet34 $\rightarrow$ WRN-28-2 (\textbf{Different architectural style}), and WRN-40-2 $\rightarrow$ WRN-40-2 (\textbf{Self}). The manual feature links are not changed once they are selected, but AFD are adaptively selected during the distillation.}
    \label{fig:attention_epoch}
\end{figure*}

\begin{table}
\centering
\begin{tabular}{c|cc|c}
\toprule
        & \multicolumn{2}{c|}{Tiny ImageNet}         & ImageNet \\ \midrule
Teacher & ResNet34        & ResNet34        & ResNet34 \\
Student & ResNet18        & ResNet34        & ResNet18 \\ \midrule
Teacher & 0.6750          & 0.6750          & 0.7355   \\
Student & 0.6530          & 0.6750          & 0.7028   \\ \midrule
KD      & 0.6818          & 0.6971          & 0.7066   \\
FitNet  & 0.6779          & 0.6892          & 0.7085   \\
ATT     & 0.6782          & 0.6961          & 0.7093   \\
RKD     & 0.6772          & 0.6846          & 0.7137   \\
CRD     & 0.6819          & 0.6968          & 0.7135   \\ \midrule
Ours    & \textbf{0.6880} & \textbf{0.6981} & \textbf{0.7138}       \\ \bottomrule 
\end{tabular}
\vspace{0.3em}
\caption{Performance comparison on large-scale datasets; Tiny ImageNet and ImageNet.}
\label{table:imagenet}
\end{table}

For tinyImageNet, we pad the images to $72 \times 72$ and then randomly cropped to $64 \times 64$ and flipped for data augmentation. We set the batch size as 128 and the maximum iteration as 200 epochs. All models are trained with stochastic gradient descent with momentum 0.9. We set weight decay as $5\times 10^{-4}$, initial learning rate as 0.1, and we divide the learning rate by 5 at 60, 120, 150 and 180 epochs. We adopt ResNet34 for the teacher and utilize ResNet34 and ResNet18 both for the student. Unlike the ResNet34 and ResNet18 architecture for large image classification, we resize the first convolutional filter size from 7 to 3. For ImageNet, we randomly crop size of $224\times 224$ of each of images and flipped for data augmentation. We optimize the student with initial learning rate as 0.1, divide it by 10 at 30, 60, 90 epochs, and set the maximum iteration as 100 epochs. We set weight decay as $10^{-4}$ and the batch size as 256. We utilize ResNet34 for the teacher and ResNet18 for the student. For tinyImageNet and ImageNet, we set the hyperparameter, $\beta$, as 50.

Table \ref{table:imagenet} shows the experiment results on tinyImageNet and ImageNet. The proposed method achieve better performance on large-scale datasets than other baseline knowledge distillation methods. It should be noted that we do not conduct L2T on these large-scaled datasets due to the heavy time complexity to update their meta-network. In addition, our method shows constantly better performances than ATT that holds the same distillation loss for the manually selected feature pairs.

\subsection{Transfer Learning}

Transfer learning with knowledge distillation utilizes the teacher pre-trained on a source domain task to train the student for a target domain task. We investigate the effectiveness of our method on transfer learning tasks. We adopt a ResNet34 pre-trained on ImageNet as the teacher network and transfer its knowledge into the students for four tasks, such as Caltech-UCSD Bird (CUB 200) \cite{cub}, MIT Indoor Scene Recognition (MIT67) \cite{mit67}, Stanford 40 Actions (Stanford40) \cite{stan40} and Stanford Dogs \cite{stan_dog}. CUB 200 consists of 5k training and 6k validation images with 200 bird species. MIT67 contains 5k training and 1k validation images with 67 type of indoor scenes. Stanford40 has 4k training and 5k validation images with 40 human actions. Stanford Dogs consists of 12k training and 8k validation images with 120 dog species. The images of transfer learning datasets consists of large scale images. The student architecture for the specific target tasks are set as ResNet18. All experiment settings are followed by those of L2T~\cite{l2t} and our hyper-parameter, $\beta$, is set as 1,000.

\begin{table}[h]
\centering
\begin{tabular}{c|cccc}
\toprule
        & \small{CUB200} & \small{MIT67}  & \small{Stanford40} & \small{Stanford Dogs} \\ \midrule
Scratch & 0.4215 & 0.4891 & 0.3693     & 0.5808        \\ 
FitNet  & 0.4893 & 0.5488 & 0.4450     & 0.6725        \\
ATT     & 0.5774 & 0.5918 & 0.5929     & 0.6970        \\
L2T     & 0.6505 & 0.6485 & 0.6308     & \textbf{0.7808}        \\ \midrule
Ours    & \textbf{0.6829} & \textbf{0.6647} & \textbf{0.6792}     & 0.7606             \\ \bottomrule 
\end{tabular}
\vspace{0.3em}
\caption{Performance comparison on multi-domain transfer learning tasks; from a ResNet34 model pre-trained with ImageNet (\textbf{source domain}) to a ResNet18 models for CUB200, MIT67, Stanford40, and Stanford Dogs (\textbf{target domains}). Scratch, ATT and L2T are referred from \cite{l2t}. All experiments are repeated 3 times.}
\label{table:multi-domain}
\end{table}

Table \ref{table:multi-domain} summarizes the experiment results for the multi-domain transfer learning tasks. For most datasets, our method shows better performance than L2T and ATT. In transfer learning, dataset from target tasks give limited information. Therefore, training without any knowledge transfer (scratch) shows the worst performance with large margin. Comparing ATT with L2T and our method, it can be seen that identifying the feature links is effective in transfer learning tasks.

\subsection{Qualitative Studies on Feature Attention}

\subsubsection{Feature Links.}
Here, we analyze the attention values, $\alpha$, learned from our proposed method in order to provide its further properties. First, we observe the attention maps of various architecture pairs, such as the same architectural style, different architectural style and identical architecture (self).

Figure \ref{fig:attention_epoch} shows the feature links $\alpha$ during the training phases. As can be seen, the traditional knowledge distillation methods manually link the teacher and student features and transfer teacher's knowledge only through the pre-defined links. In contrast to the manual feature links, our method identifies feature links in a data-driven way and thus the feature links $\alpha$ are changed over training steps and converged at the end. In the case of the same architectural style, low and mid-levels of the teacher features are linked with a low-level student feature. This indicates our method affects the student to use more layers to learn high-level of the teacher features. 
In the case of the different architectural style, the lowest-level and the highest-level features are linked among themselves and mid-level features are smoothly connected.
These results show that it is difficult to manually create feature links in different architecture styles.
In the case of the identical architectures, the features are linked in the order of the levels but the student tends to use more layers to extract high-level features. 
Based on observations, we can infer that AFD tends to link features in the same level.
However, when the teacher and student have different architectures, the connection is extended to other levels.
It is an advantage of our method that the feature links can be changed and extended according to the difference between the teacher and the student's architecture, and improves the distillation performance in various architecture settings.

\begin{figure}[h]
    \centering
    \includegraphics[width=1\columnwidth]{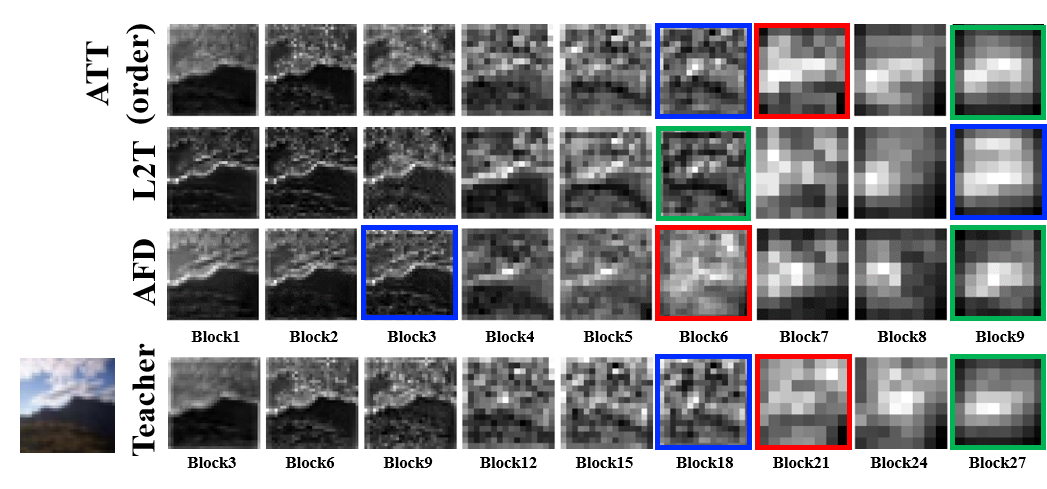}
    \caption{Activation map corresponding to each distillation methods and the teacher. The teacher is ResNet56 and the students are ResNet20. The colored boxes indicate the links from the teacher to the student.}
    \label{fig:activation_map}
\end{figure}

\subsubsection{Activation Map}
Figure~\ref{fig:activation_map} shows activation maps of a teacher and a students trained with knowledge distillation methods such as ATT with manual feature pairs, L2T, and ours. When comparing links between the teacher and student features, ATT has manually set ordered links through the levels of the features. In the L2T, the identified links switch the order of the levels; the last feature of the teacher is connected with the mid-level feature of the student (green box), and the mid-level feature of the teacher is linked to the last feature of the student (blue box). 
The identified links of the L2T may hinder the student training by transferring different order of features learned by the teacher, see green and blue boxes.
In contrast to both, our proposed method spreads the high-level features of the teacher to various feature levels of the student (blue, red, and green boxes) and identify links for the student to train while keeping the order of the features from the teacher.
More interestingly, compared with ATT, the low-level and the mid-level features of the AFD student tends to mimic activated regions of the high-level features of the teacher. 

\subsection{Sensitivity Analysis}
\begin{figure}[h]
    \centering
    \includegraphics[width=1\columnwidth]{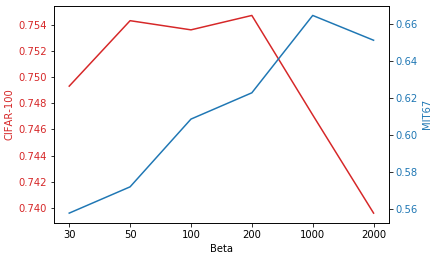}
    \caption{Sensitivity analysis for $\beta$. Red line indicates accuracy of model compression task (WRN-40-2 $\rightarrow$ WRN-16-2). Blue line indicates accuracy of transfer learning task (ResNet34 $\rightarrow$ ResNet18)}
    \label{fig:sensitivity}
\end{figure}

To investigate the impact of the hyperparameter of AFD, $\beta$, we evaluate our model by varying the value of $\beta$. $\beta$, is used to train the attention map, $\alpha$, which determines the links of the AFD network, and to decide the degree of how much the student mimic the teacher features. We perform the sensitivity analysis for $\beta$ with the model compression and the transfer learning tasks. Figure~\ref{fig:sensitivity} shows the accuracy of each task.

For the model compression task, we use WRN-40-2 as the teacher network and WRN-16-2 as the student network with CIFAR-100 dataset. As can be seen from the red line in Figure~\ref{fig:sensitivity}, the accuracy decreases when $\beta$ is more than 1,000 compared to the interval between 30 to 200. For the transfer learning task, we use ResNet34 as the teacher network and ResNet18 as the student network with MIT67 dataset. In contrast with model compression task, the transfer learning task shows better performance when the $\beta$ value is relatively large. However, we observe that the different result from the model compression task using the same network architecture of transfer learning setting with ImageNet dataset (we use $\beta$ as 50 for ImageNet). The rationale behind the gap of the hyperparameter lies in the degree of reliance on teachers' knowledge cased by the size of the dataset and the relevance between source and target tasks \cite{sp,vid,rkds}

\subsection{Ablation Studies}

\begin{table}[h]
\tabcolsep=0.17cm
\centering
\begin{tabular}{cc|cc}
\toprule
 \small{Teacher candi.} & \small{Linking method} & \small{ResNet56 $\rightarrow$ ResNet20}  \\ \midrule
 \multirow{3}{*}{\small{Random}}  & \small{Random link}       & 0.6945 $\pm$ 0.002             \\ 
   & \small{Ordered} & 0.6999 $\pm$ 0.003                     \\  
   & \small{AFD (co-train)}       &  \textbf{0.7105} $\pm$ 0.002                 \\
\midrule
\multirow{4}{*}{\small{Equal interval}} & \small{Random link}         & 0.6941 $\pm$ 0.005               \\
  & \small{Ordered}          & 0.7025 $\pm$ 0.002               \\
  & \small{AFD (pre-trained)}      & 0.7121 $\pm$ 0.001      \\ 
  & \small{AFD (co-train)}          & \textbf{0.7135} $\pm$ 0.002          \\ \bottomrule
\end{tabular}
\vspace{0.3em}
\caption{Ablation studies on selecting the teacher candidates and linking them to the student features. All experiments are repeated 5 times.}
\label{table:2+3}
\end{table}

\subsubsection{Linking methods.}
In order to reveal the benefits of the similarity-based links, we compare ours from other linking methods in Table~\ref{table:2+3}. In this experiment, we set the numbers of the teacher and student candidates same as 9 for the same architectural style (ResNet56$\rightarrow$ReNet20) to use all possible student features. We choose the teacher candidates in two ways; ``Random'' that randomly selects the candidates from all features upon all residual blocks and ``Equal interval'' that selects the features that are sequentially equidistant between themselves. For the linking method, we evaluate three linking methods; ``Random link'', ``Ordered'', ``AFD''.  

As shown in Table~\ref{table:2+3}, ``Ordered'' shows better performance than ``Random link''. It should be noted that the combination of ``Equal interval'' and ``Ordered'' is usually used when manually selecting links between the teacher and student features. Interestingly, when applying the links identified from the pre-trained AFD, we observe the performance improvement although we only change links from the usually link setting. The result proves that there is more effective way to set the links than manually decided links. 
In addition, when training AFD together, AFD shows the best performances. 
This experiments prove the superiority of AFD on identifying links of the feature pairs and transferring teacher's knowledge to the student.

\begin{table}[h]
\tabcolsep=0.2cm
\begin{subtable}{.48\textwidth}
\centering
\begin{tabular}{c|ccc}
\toprule
                          \# of candi.             & \multicolumn{3}{c}{\# of candi.(\textbf{student})} \\ \cmidrule{2-4} 
(\textbf{teacher}) & 3           & 6          & 9          \\ \midrule
3                                     & \textbf{0.7140}      & 0.7097     & 0.7115     \\
9                                     & 0.7109      & 0.7149     & 0.7135     \\
27                                    & 0.7121      & \textbf{0.7152}     & \textbf{0.7153}   \\ \bottomrule  
\end{tabular}
\caption{ResNet56 $\rightarrow$ ResNet20}
\end{subtable}
\hspace{1em}
\begin{subtable}{.48\textwidth}
\centering
\begin{tabular}{c|ccc}
\toprule
                          \# of candi.             & \multicolumn{3}{c}{\# of candi.(\textbf{student})} \\ \cmidrule{2-4} 
 (\textbf{teacher})  & 3           & 6          & 12         \\ \midrule
4                                     & 0.7698      & 0.7704     & 0.7730     \\
8                                     & \textbf{0.7714}      & 0.7717     & 0.7724     \\
16                                    & 0.7706      & \textbf{0.7733}     & \textbf{0.7747}    \\ \bottomrule  
\end{tabular}
\caption{ResNet34 $\rightarrow$ WRN-28-2}
\end{subtable}
\caption{Ablation studies on the numbers of the candidates for both the teacher and the student. The candidates are set as the output features of the residual blocks that are sequentially equidistant between themselves.}
\label{table:ablation_layer}
\end{table}

\subsubsection{Number of candidates.}
Here, we analyze the impact of the numbers of the teacher and student candidates. For this experiment, the candidates are set as the output features of the residual blocks that are sequentially equidistant between themselves. Table~\ref{table:ablation_layer} shows distillation performances over varying numbers of the candidates. When the student candidates are more than half of the total features, and using all teacher features (27 for ResNet56 and 16 for ResNet34) provides better student performances. In the other hand, with the small number of the student candidates, using all teacher features causes a information bottleneck and degrades the performances. This experiments provide a guidance to choose the number of the candidates.

\subsubsection{Distance Metrics and Pooling Methods}
We use L2 distance for distilling the teacher's feature to the student according to the trained link $\alpha$, see equation~\ref{eq:sad_loss}. However, other distance metric can used for distillation \cite{nst,vid,atts} and it may affect the behavior of the student. Therfore, we explore four distance metrics, L1, L2, KL divergence, and cosine similarity, on model compression task with WRN-40-2 as the teacher network and WRN-16-2 as the student network. Table \ref{table:distance_metric} shows that L2 distance is the optimal metric for our experiments, so we use L2 distance for the whole experiments.

Also, channel-wise pooling method, $\tilde{\phi}^{C}$, applied to feature in equation~\ref{eq:sad_loss} may affect the performance of distillation. Therefore, we compare three channel-wise pooling methods including max-pooling ($\text{max}_i |h_i|$) and average-pooling ($\frac{1}{d}\Sigma_i |h_i|^p$). We denote A1 and A2 as average pooling methods with $p=1$ and $p=2$, respectively. As can be seen in the table \ref{table:pooling}, A2 shows the best performance. Therefore, we use A2 for all experiment in this paper.

\begin{table}[h]
\centering
\begin{tabular}{cccc}
\toprule
L1     & L2              & KL     & Cosine \\ \midrule
0.7513 & \textbf{0.7547} & 0.7523 & 0.7541 \\ \bottomrule
\end{tabular}
\caption{Accuracy according to distance metrics.}
\label{table:distance_metric}
\end{table}
\vspace{-1em}

\begin{table}[h]
\centering
\begin{tabular}{ccc}
\toprule
A1     & A2              & Max      \\ \midrule
0.7520 & \textbf{0.7547} & 0.7528  \\ \bottomrule
\end{tabular}
\caption{Accuracy according to pooling methods.}
\label{table:pooling}
\end{table}

\section{Conclusion}

In this paper, we have proposed an attention-based distillation method adaptively transferring knowledge of teacher features to multiple levels of the student layers.
With the proposed method, the teacher features are linked with the student features with the attention map and the student learns from the teacher through the identified links. 
The proposed method is efficiently learned simultaneously during the student's training phase while the previous feature linking method requires an additional inner-loop procedure. 
Our experiment proves the benefits of the proposed method on two knowledge distillation applications such as model compression and transfer learning. Our further analysis shows that our method adjusts the feature levels of the student regardless the architectural styles of the teacher and student and provides better performance than the baseline methods. 

\section{Appendix A. Changes made since NeurIPS Submission}
To reflect questions and comments of reviewers of the NeurIPS, we have revised the NeurIPS submission paper. First of all, we fix the typos and notation errors raised by reviewers of NeurIPS. Second, we conduct the sensitivity analysis for the hyperparameter $\beta$ of AFD. We add the experiment result and discussion in \textit{Sensitivity Analysis} section. Third, we analyze the difference between distance metric and channel-wise pooling methods in equation 3 of the paper to search the optimal method. The details are described in \textit{Distance Metric and Pooling Methods} section.

There are some other questions raised by reviewers of NeurIPS, but we do not attach results in the paper. Our proposed method, AFD, improves the performance of existing feature distillation method ATT \cite{atts}. Feature distillation is compatible with distillation method for penultimate layer, \textit{e.g.} CRD, RKD. Therefore, we conducted an experiment for combination of CRD and AFD. The combined distillation achieves 0.44pp accuracy improvement over AFD only model. However, this experiment is somewhat orthogonal to our work, so we do not add the experiment result of combination of CRD and AFD setting in the paper.

Furthermore, one of reviewer had some question about fine-tuning experiment for transfer learning task. We add this experiment in supplementary material \textit{Appendix C. Fine-Tuning Approach for Transfer Learning} section. The fine-tuning approach is also important, but we think that the results of maintaining the same experimental conditions as designed by the L2T authors are more important. Therefore we report the results of the L2T experiment setting in the paper, and the fine-tuning results are presented in the supplementary material.

\section{Appendix B. Implementation Details of Attention-based Feature Distillation}
For all experiments, we set $d$ as 128 for the dimension of the queries, keys and positional encodings in Eq. 2. Empirically, the result does not change much depending on the dimension of $d$. For scaling the student feature, $h_s^S$, we use average pooling. Also, we utilize the identity and ReLU function as the activation function of the query and key, respectively. The kernel size and stride are determined according to the size of the student feature and teacher feature. To initialize all parameters in AFD network, we use Xavier initialization. The implementation code will be open-sourced.

\section{Appendix C. Fine-Tuning Approach for Transfer Learning}
To supplement the experiment of the transfer learning task, we conduct an experiment with fine-tuning approach. We utilize the network pre-trained with ImageNet dataset. The Table~\ref{table:fine-tune} shows the accuracy of the fine-tuned model with transfer learning. First, we can see that all distillation methods still valid with fine-tuned networks with transfer learning and AFD shows the best performance. Comparing to the training from scratch, it can be seen that the difference according to the distillation method is reduced in the case of fine-tuning. However, fine-tuning approach with transfer learning is somewhat similar to the model compression task. More research is required to analyze the difference between fine-tuning and model compression task to reveal further property of pre-trained network.

\begin{figure*}[t]
    \centering
    \includegraphics[width=1.8\columnwidth]{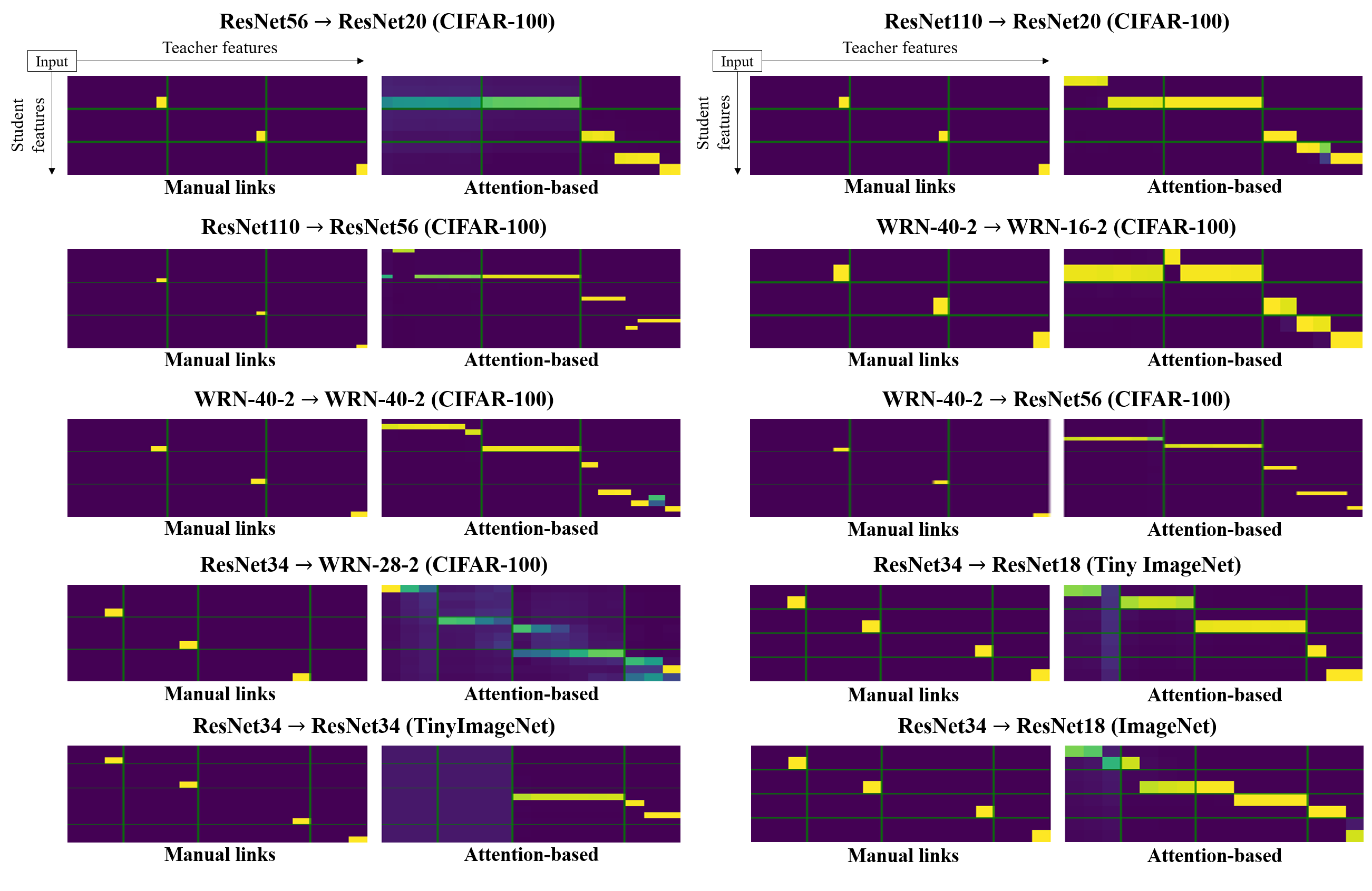}
    \caption{Manually designed feature links and attention-based feature links. Divided into green lines, if the size or channel dimension of feature is different.}
    \label{fig:appendix_a}
\end{figure*}

\begin{table}[h]
\centering
\begin{tabular}{cc}
\toprule
\multicolumn{2}{c}{ \textbf{Pretraining - Distillation}} \\ \midrule
 Fine-tuning &  0.7067 \\
 FitNet &  0.7172 \\
 ATT &  0.7254 \\
 L2T &  0.7485 \\
 AFD &  \textbf{0.7507} \\ \bottomrule
\end{tabular}
\caption{Performance comparison between transfer learning task with fine-tuning; from a ResNet34 model to a ResNet18 model for MIT67 dataset. Both models are pre-trained with ImageNet dataset.}
\label{table:fine-tune}
\end{table}

\section{Appendix D. Additional Results}
This section provides additional results of the ablation study and the qualitative studies. Table \ref{table:add_link} shows the performance according to the various linking method between the teacher and student with the different architectural style. AFD shows the best performance when the architectural style is different, such as the result of the \textit{Ablation Studies} section of the paper.

\begin{table}[h]
\small
\tabcolsep=0.17cm
\centering
\begin{tabular}{cc|c}
\toprule
 Teacher candi. & Linking method & ResNet34 $\rightarrow$ WRN-28-2 \\ \midrule
 \multirow{3}{*}{Random}  & Random link            & $0.7567 \pm 0.006$              \\ 
   & Ordered           & $0.7607 \pm 0.005$              \\  
   & AFD (co-train)            & \textbf{0.7702} $ \pm$0.004              \\
\midrule
\multirow{4}{*}{Equal interval} & Random link                 & $0.7504 \pm 0.008$           \\
  & Ordered            & $0.7544 \pm 0.003$              \\
  & AFD (pre-trained)             & 0.7723 $\pm$ 0.003   \\ 
  & AFD (co-train)           & \textbf{0.7726} $\pm$ 0.003              \\ \bottomrule
\end{tabular}
\vspace{0.3em}
\caption{Ablation studies on selecting the teacher candidates and linking them to the student features. All experiments are repeated 5 times.}
\label{table:add_link}
\end{table}

In addition, we compare manual links with converged attention-based links with various architectures. Figure \ref{fig:appendix_a} shows the feature links with various architecture pairs. This results provide the guidance to determine the feature links of various architecture pairs.  

\bibliography{ref}
\end{document}